\documentclass{article}

\usepackage[english]{babel}

\usepackage[letterpaper,top=2cm,bottom=2cm,left=3cm,right=3cm,marginparwidth=1.75cm]{geometry}
\usepackage[numbers]{natbib}
\usepackage{notoccite}

\usepackage{graphicx}
\usepackage{array} 
\usepackage{float}
\usepackage{amsmath}
\usepackage{graphicx}
\usepackage[colorlinks=true, allcolors=blue]{hyperref}

\title{Ensemble Model With Bert,Roberta and Xlnet For Molecular property prediction}
\author{Junling Hu\thanks{University of Sussex, \texttt{Email:jh2023@sussex.ac.uk}}}

\date{}

\begin{document}
\maketitle

\begin{abstract}
This paper presents a novel approach for predicting molecular properties with high accuracy without the need for extensive pre-training. Employing ensemble learning and supervised fine-tuning of BERT, RoBERTa, and XLNet, our method demonstrates significant effectiveness compared to existing advanced models. Crucially, it addresses the issue of limited computational resources faced by experimental groups, enabling them to accurately predict molecular properties. This innovation provides a cost-effective and resource-efficient solution, potentially advancing further research in the molecular domain.
\end{abstract}

\hspace{1em}Keywords:Ensemble Learning, BERT, RoBERTa, XLNet, Molecular Property Prediction

\section{Introduction}

\par
\hspace{2em}In recent years, deep learning models have rapidly expanded their application in the field of chemical science\cite{1}, including areas such as drug discovery, molecular generation, and molecular property prediction\cite{2,3,4}. Molecular properties play a pivotal role in various fields including chemistry, drug discovery, and healthcare. They connect disciplines such as quantum mechanics, physical chemistry, biophysics, and physiology. Computer-assisted methods facilitate the rapid prediction of molecular properties\cite{5}. 
\par
Traditional machine learning methods often require manual selection and construction of features for training models, such as Extended-Connectivity Fingerprints (ECFP)\cite{6}. As data volume increases, this process becomes time-consuming and may lead to performance saturation, affecting prediction accuracy\cite{7}. Currently, various deep learning models have been proposed for predicting molecular properties. Common methods for molecular property prediction are based on one-dimensional sequences\cite{8,9,10}, such as SMILES, and two-dimensional representations\cite{11,12,13}, such as molecular graphs. 
\par
With the introduction of the Transformer model\cite{14} in 2017 and Google's
subsequent release of the BERT model in 2018, pre-training strategies have achieved great success in the field of natural language processing (NLP). Models based on SMILES sequences can be pre-trained using specially designed tasks such as the masked language model\cite{15,16}, focusing more on the context of molecular sequences. Graph models typically capture structural features by constructing a graphical
representation of molecules\cite{17,18}, where nodes represent atoms and edges represent chemical bonds between atoms. Randomly masking parts of a graph and training the model to predict the attributes or relationships of these masked parts is a key aspect of self-supervised learning. Regardless of the molecular representation chosen, pre-training the input model is crucial\cite{19}. However, pre-training models usually require substantial computational resources and large datasets, which is very time-consuming\cite{20}. This presents significant challenges for environments with limited computational resources\cite{21}. Additionally, pre-training may lead to excessive optimization for specific tasks, limiting its applicability in processing small-scale or specific types of datasets. Therefore, this study explores a strategy of fully fine-tuning these pre-trained architectural models from a state of random initialization, abandoning the traditional pre-training steps. To
overcome the limitations of individual models and compensate for the lack of pre-training, this paper adopts an ensemble learning method\cite{22,23} using BERT\cite{24}, RoBERTa\cite{25}, and XLNet\cite{26} models for predicting molecular properties. Ensemble learning combines multiple models, utilizing their different perspectives and strengths
to improve prediction accuracy. By stacking the outputs of multiple models, the ensemble method can compensate for the weaknesses of individual models, providing more robust and accurate property predictions. BiLSTM\cite{27} is used as a component of the base predictor, and BaggingRegressor is used as a meta-predictor for the final prediction. 
\par
In light of this, the study aims to develop a molecular property prediction scheme that reduces reliance computational resources without significantly sacrificing prediction accuracy. This approach is intended to provide a feasible solution for environments with limited computational resources. The study hopes to demonstrate that effective molecular property prediction can still be achieved without large-scale
pre-training, potentially rivaling the performance of current advanced models.

\section{Relevant Work}
\hspace{2em}Tokenization is a crucial preprocessing step in NLP, significantly impacting the quality of predictions. Therefore, the first key issue is how to represent molecules. With the rapid development of NLP models, particularly those in the Transformer family, tokenizers can encode words or sentences. This allows the conversion of one-dimensional SMILES information into a tokenized language understandable by machines. Tokenizers use byte-pair encoding to construct the vocabulary for model inputs. Choosing a one-dimensional approach like SMILES as the input for NLP models has significant advantages compared to two-dimensional methods. SMILES\cite{28} is a character encoding system used to represent chemical molecules. It transforms complex molecular structures into one-dimensional string representations by sequentially depicting atoms within the molecule. This transformation is achieved by applying a depth-first search algorithm to the molecular graph, generating a linear character sequence that reflects the molecular structure. This approach not only simplifies the model's processing flow but also significantly reduces computational complexity. Due to its structural similarity to sentences in human language, the SMILES format enhances data interpretability, allowing NLP methods to be effectively applied in chemical data analysis. Hence, the SMILES format enables deep learning-based models to more effectively capture fundamental molecular features and generate accurate molecular property predictions. 
\par
However, previous studies have indicated limitations in SMILES representation. Different carbon atoms in a molecule may have different relationships with other atoms and occupy different positions, potentially corresponding to different properties. In SMILES, atoms of the same element with different properties are represented in the same way. Therefore, relying solely on SMILES for molecular property prediction is inaccurate. This problem is viewed as a challenge, prompting researchers to develop new SMILES representations to overcome the deficiencies of traditional representations\cite{29}. DeepSMILES\cite{30} increases the probability of generating valid molecules by introducing closing brackets or single symbols at cyclic positions. SELFIES\cite{31} proposes a different molecular representation based on Chomsky
Type-2 grammar, introducing a grammar-based molecular representation framework significantly different from traditional SMILES. 
\par
Furthermore, Ucak et al. introduced the Atom in SMILES (AIS) method\cite{32}, eliminating ambiguity in property generation from SMILES representations. This formalized AIS description provides comprehensive atomic and environmental details, converting SMILES into nuanced atom-level representations, enhancing understanding of molecular structure and properties. From their research, AIS outperforms other SMILES tokenization methods in prediction accuracy, enabling sequence-based models to effectively utilize high-quality SMILES representations. 
\par
Selecting appropriate molecular representations and pairing them with the right models iscrucial in the field of molecular prediction. In terms of models, the rapid development of the Transformer family\cite{33} has facilitated their swift application in molecular language modeling\cite{34,35}. The Transformer outperforms traditional RNN models in terms of performance, becoming the most versatile model to date. Innovations and improvements upon the Transformer framework in models like BERT, RoBERTa, and XLNet have led to superior performance in handling complex language modeling tasks. 
\par
In the evolving field of molecular property prediction, the adoption of NLP techniques signifies a leap forward. Ross et al. pioneered with MolFormer\cite{36}, harnessing over 1.1 billion molecules to forecast chemical behaviors, showcasing the transformer model's prowess in capturing intricate molecular details. This approach underlines the potential of large-scale molecular language models in scientific discovery. 
\par
Parallel to Born et al. introduced the Regress Transformer (RT)\cite{37}, a novel concoction blending regression analysis with conditional generation tasks. Using the XLNet architecture, RT has surpassed existing models in both chemical and protein language modeling, illustrating the vast potential of combining numeric and text tokens for molecular science. Ross et al., Wang et al. introduced SMILES-BERT\cite{38}, a model predicated on unsupervised pretraining that has demonstrated remarkable predictive accuracy across
several benchmarks. Its success on datasets like QM9 and ESOL highlights the model's ability to decipher complex chemical information, positioning it as a cornerstone for future explorations in drug discovery and material sciences. 
\par
Furthering this trajectory, Yu et al. presented SolvBERT\cite{39}, a model specifically fine-tuned for solvation properties, marking a significant advancement in understanding molecular interactions through NLP models. Similarly, Li et al. developed Mol-BERT\cite{40}, leveraging a vast corpus of SMILES strings to achieve unprecedented accuracy in molecular property predictions, illustrating the model's superiority in tasks that span across diverse molecular datasets. 
\par
Completing this panorama of innovation, Liu et al.\cite{41}expanded on these foundations with MolRoPE-BERT, integrating innovative position encoding methods to refine predictions further. This model's performance, validated on multiple benchmark datasets, exemplifies the continuous enhancement of molecular property prediction models.

\section{Methodology}
\subsection{Data Set}
\begin{figure}[H]
    \centering
    \includegraphics[width=1\textwidth]{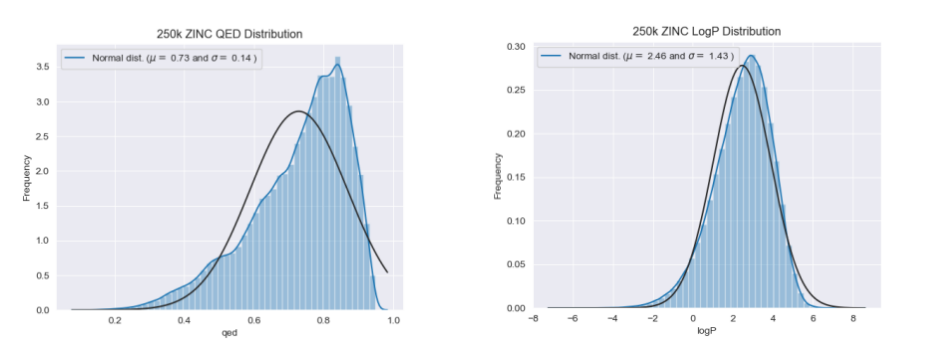}
    \caption{ZINC250k dataset qed and logP property distribution histogram}
    \label{fig1}
\end{figure}

\hspace{2em}This project maintains two datasets: zinc250k and zinc350k. Zinc250k(Fig.\ref{fig1}) is a subset of the zinc12 dataset \cite{42}, which contains 250,000 
organic molecules. Each molecule is provided with a SMILES and two properties. 
The dataset includes real values for the log octanol-water partition coefficient (logP), 
which is a measure of lipophilicity and indicates how hydrophobic a compound is. 
Additionally, each molecule is scored with a quantitative estimate of drug-likeness 
(QED), which reflects the molecule's potential to be a drug based on its 
physicochemical properties; QED values in this dataset range from 0.11 to 0.95.

\begin{figure}[H]
    \centering
    \includegraphics[width=1\textwidth]{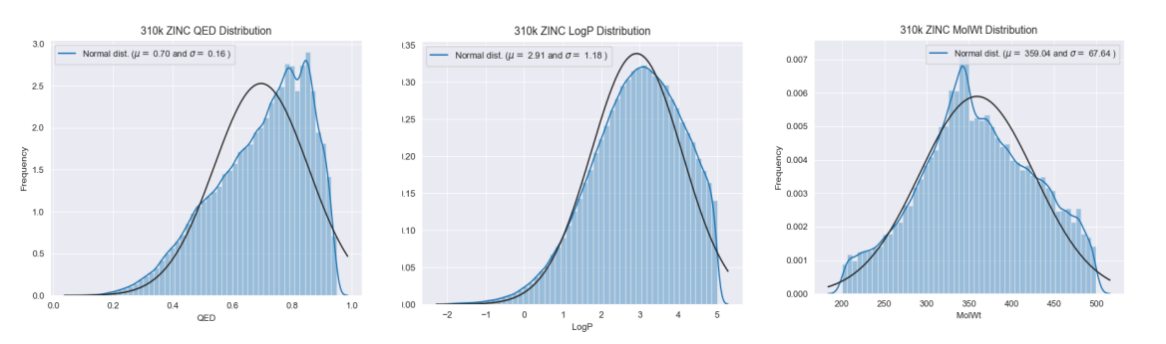}
    \caption{ZINC310k dataset qed, logP and MolWt property distribution histogram}
    \label{fig2}
\end{figure}

Zinc310k(Fig.\ref{fig2}) is a derivative of the complete zinc15 dataset \cite{43}, featuring 
310,000 molecules. Similar to Zinc250k, each entry in this dataset is associated with a 
SMILES representation, QED score, and logP value. Furthermore, this dataset 
includes the molecular weight (MW) of each molecule, which is the mass of a single 
molecule of a substance and is typically measured in atomic mass units (g/mol). In 
Zinc310k, QED values span from 0.07 to 0.94, logP values are between -1.99 to 4.99, 
and MW ranges from 200 to 500 g/mol, providing a diverse set of molecules for the 
analysis of physicochemical and structural properties relevant to drug discovery.

\subsection{Data Preprocessing}
\begin{figure}[H]
    \centering
    \includegraphics[width=1\textwidth]{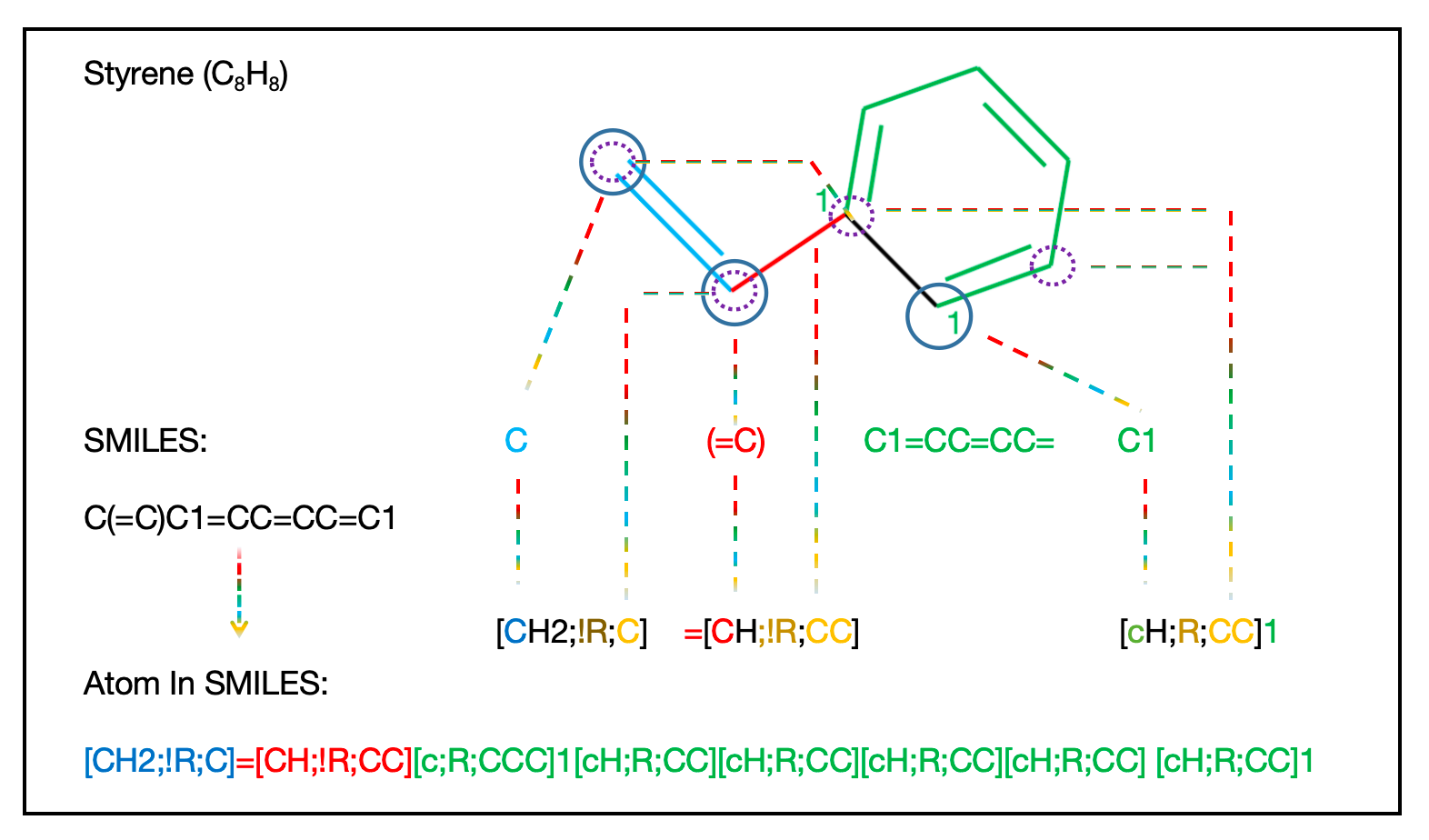}
    \caption{An example illustrates the SMILES expression of the Styrene molecule (C(=C)C1=CC=CC=C1) and the step-by-step transformation process into AIS.}
    \label{fig3}
\end{figure}

\hspace{2em}Our model selects AIS as the input and converts SMILES into AIS representation(Fig.\ref{fig3}).This conversion involves three key elements: the central atom, 
ring information, and neighboring atoms interacting with the central atom, enclosed within square brackets and separated by semicolons.The central atom's representation includes details about the corresponding SMILES atom, along with the count of neighboring hydrogen atoms. !R denotes the atom's exclusion from a ring, whereas R signals its inclusion in a ring.In cases where an atom is part of a benzene ring, its representation employs lowercase letters to indicate aromaticity. The final element of AIS illustrates atoms adjacent to the central atom. AIS ensures a direct mapping of represented atoms to those in the original SMILES, maintaining consistency with non-atomic symbols. Chirality information can be attached to the central atom using @ or @@ suffixes.

\begin{figure}[H]
    \centering
    \includegraphics[width=0.7\textwidth]{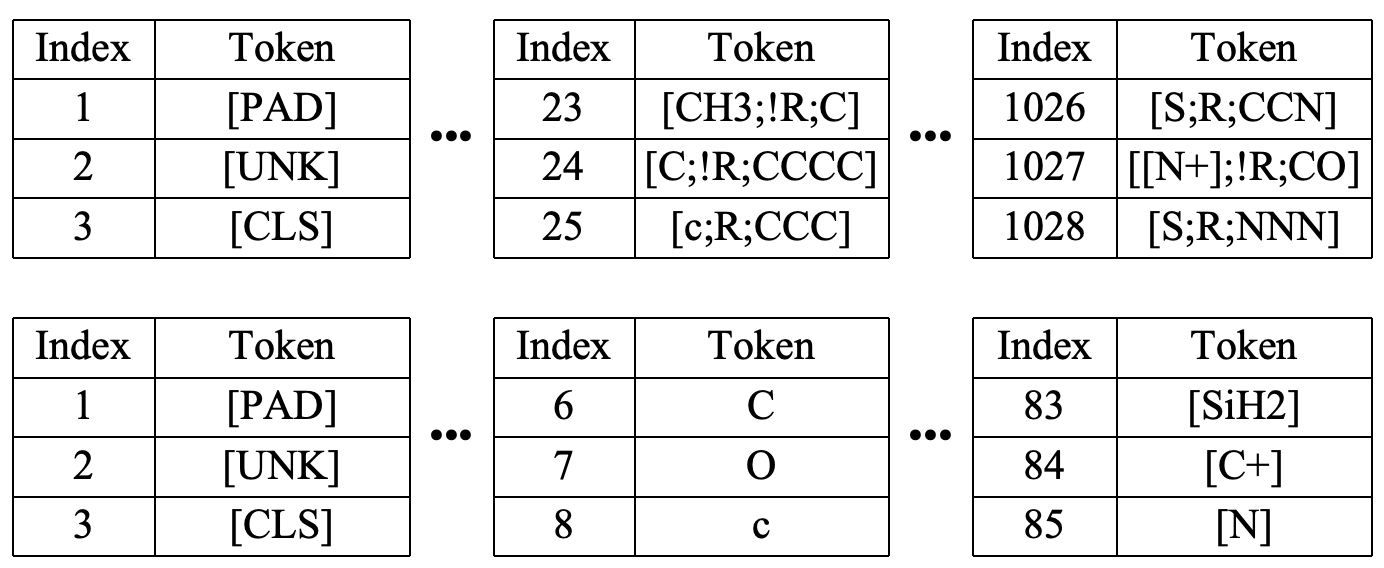}
    \caption{The vocabularies for AIS(Up) and SMILES(Down) were created based on the zinc250k and zinc310k datasets}
    \label{fig4}
\end{figure}

This formalized AIS representation offers a simplified and systematic approach to address the limitations of traditional SMILES tokenization, enhancing the predictive quality of molecular property prediction tasks.
After converting SMILES into Atom In SMILES (AIS), we treat each individual atom and special symbol as a separate token to construct the vocabulary(Fig.\ref{fig4}). When analyzing two datasets, we observed that traditional SMILES representation covered approximately 80 different tokens, while AIS representation increased to around 1000 tokens. This significant change highlights the richness and depth of AIS in capturing molecular structural details, making SMILES expression more akin to human language. AIS, with its expanded vocabulary, can provide a more detailed description of molecular properties, thereby offering deeper insights to deep learning models.

\subsection{Ensemble Model}
\begin{figure}[H]
    \centering
    \includegraphics[width=1\textwidth]{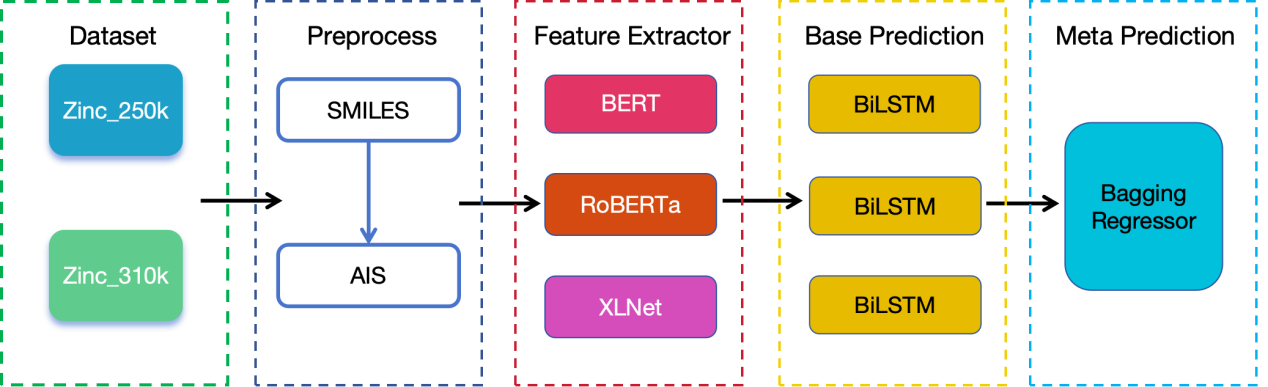}
    \caption{The structure of ensemble model}
    \label{fig5}
\end{figure}

\hspace{2em}We designed an ensemble model(Fig.\ref{fig5}) to predict molecular properties, incorporating BERT, RoBERTa, and XLNet as feature extractors to obtain AIS text features. These features are then passed to the base predictor, BiLSTM, which is an integral part of the base predictor, making full use of its sensitivity to sequential data and strong time-feature capturing capabilities. By integrating information in both forward and backward directions, the results are ultimately aggregated to the meta-learner (BaggingRegressor) to derive the final prediction. BaggingRegressor is a powerful ensemble learning algorithm used for regression tasks. It is based on the Bagging (Bootstrap Aggregating) principle, which improves overall performance by combining predictions from multiple base learners. 

To optimize training speed and efficiency, the layers of BERT and XLNet were reduced to six layers, while a small version of RoBERTa was employed. The table 1 below details the specific parameters of the model.
\begin{table}[H]
\centering
\caption{Parameters set in model}
\begin{tabular}{cc}
\hline
Hyperparameter Name & Value   \\ \hline
Learning Rate       & 0.00001 \\
Batch Size          & 16      \\
Dropout             & 0.1     \\
Hidden Size         & 768     \\
Attention Heads     & 12      \\
Epochs              & 100     \\
Loss Function       & MAE     \\ \hline
\end{tabular}
\vspace{0.2cm} 
\end{table}

\subsection{Baseline Methods}
\hspace{2em}We compare our ensemble model with the current state-of-the-art baseline models, which include:
ASVAE (All SMILES Variational Autoencoder)\cite{44}: ASVAE deploys a diversified SMILES representation and employs a stack of recursive neural networks to encode individual molecules. By combining semi-supervised and fully supervised learning, it significantly improves the accuracy of predicting various molecular properties. ASVAE has demonstrated its effectiveness in predicting attributes such as logP, molecular weight, and drug likeness (QED) in prior research, surpassing other advanced models \cite{45,46,47,48}. We utilize the same dataset as ASVAE and consider it as a benchmark model. Although ASVAE's research reports only the MAE metric, which we consider a critical indicator as it directly reflects the average absolute difference between predicted and actual values, we specifically focus on MAE during the comparison.

GROVER(Graph Representation frOm self-superVised mEssage passing tRansformer)\cite{49}. GROVER is a hybrid graph neural network (GNN) and Transformer architecture that takes molecular graphs as input, aiming to optimize the graph representation of molecules. What makes this model unique is its combination of two GNN Transformers—one tailored for nodes and the other for edges. These components are structurally similar but handle different features. Notably, one GNN component is specifically designed to transform graph information into features required by the Transformer. GROVER employs transfer learning techniques, similar to other sequence-based models, to improve the training efficiency and accuracy of downstream tasks. In our study, we used the pre-trained GROVER model and fine-tuned it on our dataset for performance comparison.

CHEM-BERT\cite{50}. Exploiting SMILES strings as input, CHEM-BERT focuses on learning SMILES features during BERT pre-training. Additionally, this method incorporates matrix embedding layers for structural learning and quantitative estimation of drug similarity (QED) prediction tasks. Their approach has shown outstanding performance on multiple benchmark datasets, demonstrating its effectiveness in generalizing molecular data and improving downstream task prediction accuracy. To compare the performance of different models, we conducted a similar fine-tuning process on CHEM-BERT.

D-MPNN(Directed Message Passing Neural Network)[49]\cite{51}. D-MPNN takes molecular graphs as input and innovatively uses directed edges instead of atoms to convey information within the neural network structure, enhancing the molecular representation learning process. It can directly and efficiently handle molecular structural data without the need for pre-training on large-scale datasets. Extensive evaluations on various public and proprietary datasets have consistently shown that D-MPNN outperforms traditional models using fixed molecular descriptors and other graph neural architectures.

\subsection{Evaluation Metrics}
\hspace{2em}This study depends on three main evaluation metrics to assess model performance:
Root Mean Square Error (RMSE), Mean Absolute Error (MAE), and Coefficient of
Determination (R²). RMSE is the square root of the average of the squared prediction errors and
reflects the volatility in the model's predictions.
\[  
\text{RMSE} = \sqrt{\frac{1}{n} \sum_{i=1}^{n} (y_i - \hat{y}_i)^2}  
\] 
\hspace{2em}MAE measures the average absolute difference between predicted and actual
values, providing an indication of the average error level.
\[  
\text{MAE} = \frac{1}{n} \sum_{i=1}^{n} |y_i - \hat{y}_i|  
\]  
\hspace{2em}R² is a statistical measure used to assess the accuracy of model predictions, with
values ranging from 0 to 1, where values closer to 1 indicate better model predictive
performance.
\[  
R^2 = 1 - \frac{\sum_{i=1}^{n} (y_i - \hat{y}_i)^2}{\sum_{i=1}^{n} (y_i - \bar{y})^2}  
\]  
\section{Experiment}
\begin{figure}
    \centering
    \includegraphics[width=1\textwidth]{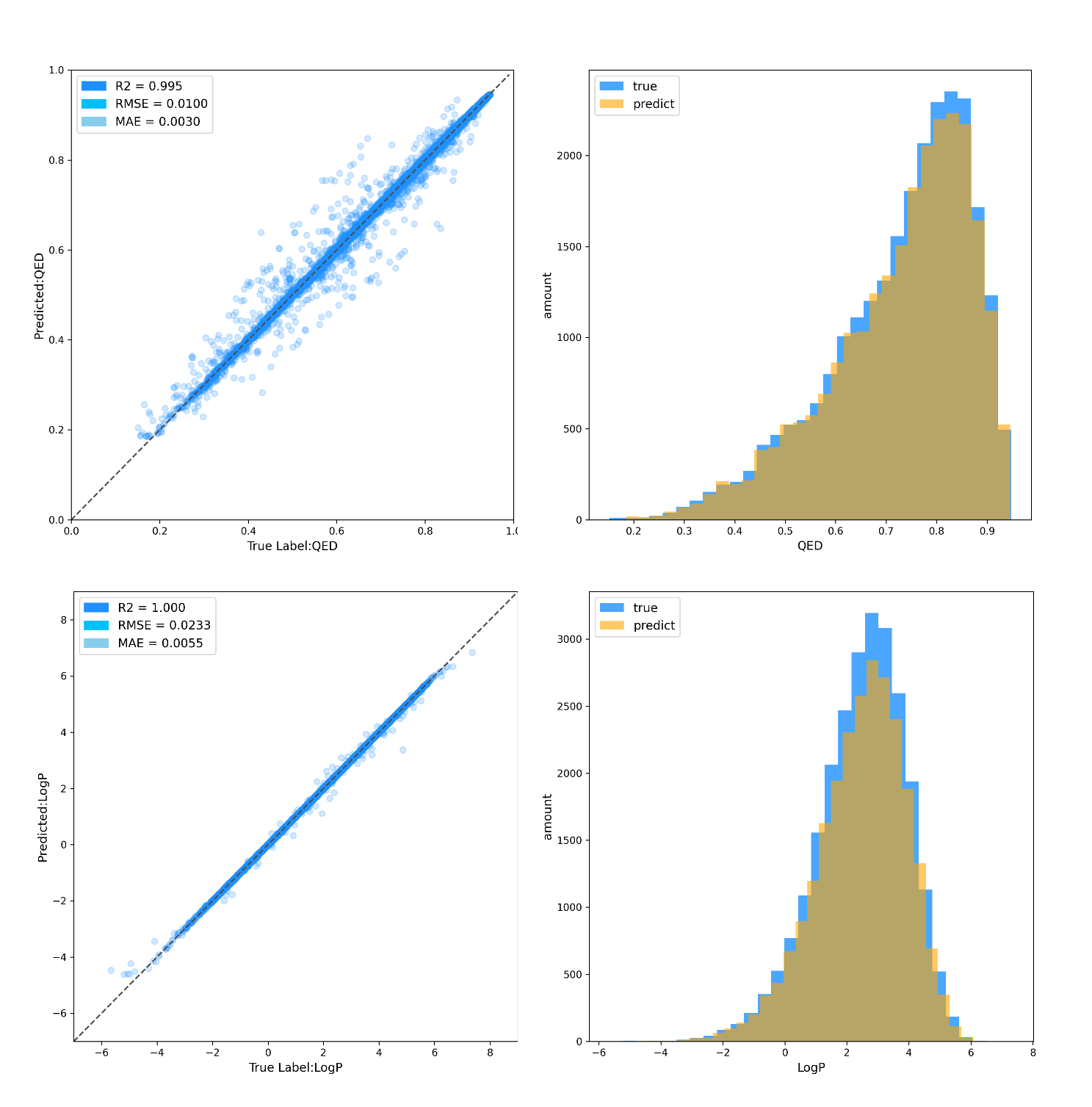}
    \caption{Regressions (left) and distributions (right) of true and predicted values of qed and logP from ZINC250k}
    \label{fig6}
\end{figure}

\begin{figure}[H]
    \centering
    \includegraphics[width=1\textwidth]{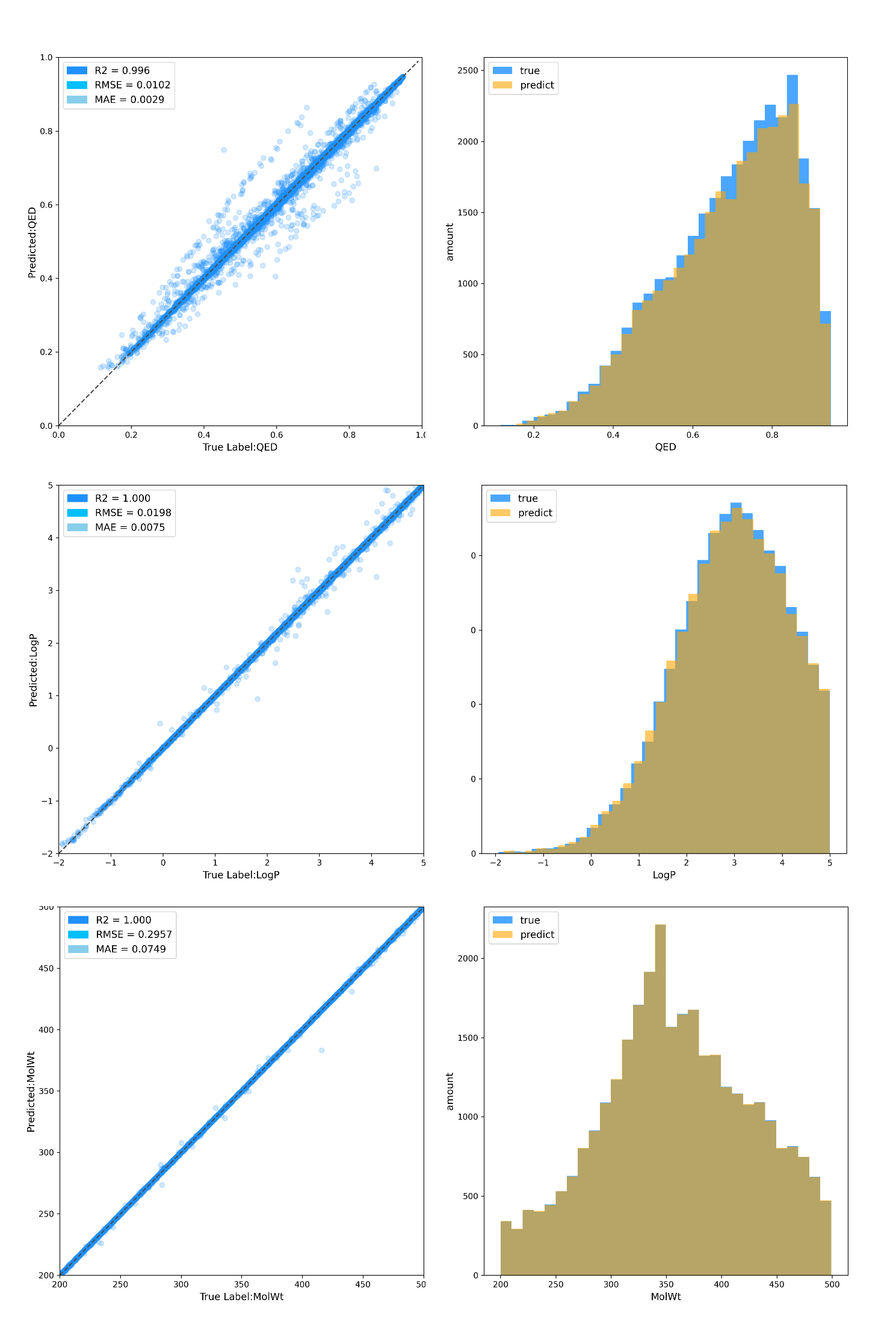}
    \caption{Regressions (left) and distributions (right) of true and predicted values of qed, logP and MolWt from
ZINC310k}
    \label{fig7}
\end{figure}

\hspace{2em}Scatter and histogram plots utilizing predicted and true values offer a direct visual
assessment of the model's performance. The final Mean Absolute Error (MAE) and
Root Mean Square Error (RMSE) across all attributes in the test set, as presented in
Fig.\ref{fig6}-\ref{fig7}, indicate errors within a tight margin of 0.5\%. Scatter plots exhibit a strong
correlation between predictions and actual values, with QED showing an R² of 0.996
and logP and MolWt achieving a perfect R² of 1.000, suggesting an almost flawless
predictive model. The histograms' alignment of true and predicted value distributions
further confirms the model's precision.
Table 2 reports the MAE, RMSE, and R² on the zinc250k dataset. From this table, we observe that our ensemble model achieves an MAE of 0.003 on the QED property, which is 42\% better than the baseline model ASVAE and outperforms other graph-based and sequence-based models. For the logP property, the ensemble model achieves an MAE of 0.005, which is on par with the baseline ASVAE but significantly better than other models
\begin{table}[htbp]
\centering
\caption{The performance of ZINC250k dataset}
\resizebox{\textwidth}{!}{
\begin{tabular}{lccccccccll}
\hline
               & \multicolumn{3}{c}{QED}                                                     & \multicolumn{1}{l}{} & \multicolumn{3}{c}{LogP}                                                    & \multicolumn{3}{l}{}                        \\ \cline{2-4} \cline{6-8}
Model          & \multicolumn{1}{l}{MAE} & \multicolumn{1}{l}{RMSE} & \multicolumn{1}{c}{R²} & \multicolumn{1}{l}{} & \multicolumn{1}{l}{MAE} & \multicolumn{1}{l}{RMSE} & \multicolumn{1}{c}{R²} & \multicolumn{3}{l}{Molecular representation} \\ \hline
ASVAE          & 0.0052                  & \textbackslash{}         & \textbackslash{}       &                      & 0.005                   & \textbackslash{}         & \textbackslash{}       & \multicolumn{3}{c}{SMILES}                  \\
GROVER         & 0.0056                  & 0.0083                   & 0.9964                 &                      & 0.019                   & 0.036                    & 0.9993                 & \multicolumn{3}{c}{Graphs}                  \\
D-MPNN         & 0.0056                  & 0.0094                   & 0.9953                 &                      & 0.016                   & 0.029                    & 0.9995                 & \multicolumn{3}{c}{Graphs}                  \\
CHEM-BERT      & 0.0049                  & 0.0102                   & 0.9934                 &                      & 0.011                   & 0.023                    & 0.9996                 & \multicolumn{3}{c}{SMILES}                  \\
Ensemble model & \textbf{0.0030}         & 0.0100                   & 0.9948                 &                      & \textbf{0.005}          & 0.023                    & 0.9997                 & \multicolumn{3}{c}{Atom In SMILES}          \\ \hline
\end{tabular}
}
\end{table}

Table 3 reports the MAE, RMSE, and R² on the zinc310k dataset. Here, we can
observe that our ensemble model achieves an MAE of 0.0029 for QED and 0.07 for
MolWt, improving by 54\% and 66\%, respectively, compared to the baseline ASVAE. For the logP property, it remains comparable to the baseline ASVAE. These results
further confirm the effectiveness of the ensemble model in multi-property prediction.

\begin{table}[htbp]
\centering
\caption{The performance of ZINC310k dataset}
\resizebox{\textwidth}{!}{
\begin{tabular}{lccccccccccccll}
\hline
               & \multicolumn{3}{c}{QED}                                               & \multicolumn{1}{l}{} & \multicolumn{3}{c}{LogP}                                              & \multicolumn{1}{l}{} & \multicolumn{3}{c}{MolWt}                                             & \multicolumn{3}{l}{}                        \\ \cline{2-4} \cline{6-8} \cline{10-12}
Model          & \multicolumn{1}{l}{MAE} & \multicolumn{1}{l}{RMSE} & R²               & \multicolumn{1}{l}{} & \multicolumn{1}{l}{MAE} & \multicolumn{1}{l}{RMSE} & R²               & \multicolumn{1}{l}{} & \multicolumn{1}{l}{MAE} & \multicolumn{1}{l}{RMSE} & R²               & \multicolumn{3}{l}{Molecular representation} \\ \cline{1-8} \cline{10-15} 
ASVAE          & 0.0064                  & \textbackslash{}         & \textbackslash{} &                      & 0.007                   & \textbackslash{}         & \textbackslash{} &                      & 0.21                    & \textbackslash{}         & \textbackslash{} & \multicolumn{3}{c}{SMILES}                  \\
GROVER         & 0.0052                  & 0.0073                   & 0.9977           &                      & 0.009                   & 0.018                    & 0.9997           &                      & 0.26                    & 0.35                     & 0.9999           & \multicolumn{3}{c}{Graphs}                  \\
D-MPNN         & 0.0059                  & 0.0100                   & 0.9959           &                      & 0.017                   & 0.034                    & 0.9991           &                      & 1.55                    & 3.63                     & 0.9997           & \multicolumn{3}{c}{Graphs}                  \\
CHEM-BERT      & 0.0045                  & 0.0103                   & 0.9948           &                      & 0.008                   & 0.020                    & 0.9997           &                      & 0.40                    & 0.63                     & 0.9998           & \multicolumn{3}{c}{SMILES}                  \\
Ensemble model & \textbf{0.0029}         & 0.0101                   & 0.9958           &                      & \textbf{0.007}          & 0.019           & 0.9997           &                      & \textbf{0.07}           & 0.29                     & 0.9999           & \multicolumn{3}{c}{Atom In SMILES}          \\ \hline
\end{tabular}
}
\end{table}
Furthermore, the results indicate that the type of input data significantly
influences model performance for the same dataset. Models that use sequences as
input perform better overall than models that use graphs as input. Different types of
input data, such as SMILES sequences and graph representations, have a notable
impact on model performance. In our experiments, models that use SMILES
sequences as input outperform those that use graphs as input. This may be attributed
to the richer chemical information provided by SMILES sequences, enhancing the
model's learning capacity. Our experimental results highlight the excellent performance of the ensemble
model on most performance metrics, particularly in terms of MAE. This finding
underscores the potential of ensemble methods in improving prediction accuracy. However, we also note that on certain specific properties, the ensemble model
performs comparably to single models. This suggests the need for further exploration
of optimal configurations for different model combinations in future research.

\section{Ablation Study}
\hspace{2em}Ablation Study in this project evaluates the impact of different language models
(BERT, RoBERTa, XLNet) in combination with BiLSTM and compares two different
input types (AIS and SMILES) on model performance(Table 4-8){/ref}.
\begin{table}[htbp]
\centering
\caption{The Ablation Study of ZINC250k dataset with QED property}
\resizebox{\textwidth}{!}{
\begin{tabular}{llllllllllll}
\hline
                       & \multicolumn{3}{l}{Training} &  & \multicolumn{3}{l}{Vaildation} &  & \multicolumn{3}{l}{Test} \\ \cline{2-4} \cline{6-8} \cline{10-12} 
Model                  & \multicolumn{1}{l}{MAE} & \multicolumn{1}{l}{RMSE} & \multicolumn{1}{c}{R²} & \multicolumn{1}{l}{} & \multicolumn{1}{l}{MAE} & \multicolumn{1}{l}{RMSE} & \multicolumn{1}{c}{R²} & \multicolumn{1}{l}{} & \multicolumn{1}{l}{MAE} & \multicolumn{1}{l}{RMSE} & \multicolumn{1}{c}{R²} \\ \hline
BERT+BiLSTM            & 0.0030   & 0.0080  & 0.9953  &  & 0.0037   & 0.0137   & 0.9889   &  & 0.0038 & 0.0144 & 0.9870  \\[\extrarowheight] 
RoBERTa+BiLSTM         & 0.0035   & 0.0081  & 0.9947  &  & 0.0040   & 0.0130   & 0.9899   &  & 0.0039 & 0.0127 & 0.9899 \\
XLNet+BiLSTM           & 0.0030   & 0.0063  & 0.9974  &  & 0.0038   & 0.0108   & 0.9899   &  & 0.0039 & 0.0114 & 0.9918  \\
BiLSTM                 & 0.0054   & 0.0152  & 0.9860  &  & 0.0068   & 0.0186   & 0.9795   &  & 0.0070 & 0.0194 & 0.9765  \\
BERT+BiLSTM(SMILES)    & 0.0035   & 0.0090  & 0.9949  &  & 0.0042   & 0.0141   & 0.9882   &  & 0.0041 & 0.0142 & 0.9874  \\
RoBERTa+BiLSTM(SMILES) & 0.0035   & 0.0083  & 0.9957  &  & 0.0040   & 0.0121   & 0.9913   &  & 0.0041 & 0.0126 & 0.9904 \\
XLNet+BiLSTM(SMILES)   & 0.0038   & 0.0092  & 0.9947  &  & 0.0045   & 0.0127   & 0.9904   &  & 0.0044 & 0.0127 & 0.9903 \\
BiLSTM(SMILES)         & 0.0107   & 0.0221  & 0.9701  &  & 0.0120   & 0.0250   & 0.9620   &  & 0.0118 & 0.0246 & 0.9627  \\
Ensemble(SMILES)       &          &         &         &  &          &          &          &  & 0.0035 & 0.0101 & 0.9958 \\ \hline
\end{tabular}
}
\end{table}
\begin{table}[htbp]
\centering
\caption{The Ablation Study of ZINC250k dataset with LogP property}
\resizebox{\textwidth}{!}{
\begin{tabular}{llllllllllll}
\hline
                       & \multicolumn{3}{l}{Training} &  & \multicolumn{3}{l}{Vaildation} &  & \multicolumn{3}{l}{Test} \\ \cline{2-4} \cline{6-8} \cline{10-12} 
Model                  & \multicolumn{1}{l}{MAE} & \multicolumn{1}{l}{RMSE} & \multicolumn{1}{c}{R²} & \multicolumn{1}{l}{} & \multicolumn{1}{l}{MAE} & \multicolumn{1}{l}{RMSE} & \multicolumn{1}{c}{R²} & \multicolumn{1}{l}{} & \multicolumn{1}{l}{MAE} & \multicolumn{1}{l}{RMSE} & \multicolumn{1}{c}{R²} \\ \hline

BERT+BiLSTM            & 0.0077   & 0.0103  & 0.9999  &  & 0.0085   & 0.0244   & 0.9996   &  & 0.0087 & 0.0268 & 0.9996 \\
RoBERTa+BiLSTM         & 0.0174   & 0.0255  & 0.9996  &  & 0.0186   & 0.0366   & 0.9993   &  & 0.0186 & 0.0346 & 0.9994 \\
XLNet+BiLSTM           & 0.0190   & 0.0248  & 0.9996  &  & 0.0233   & 0.0412   & 0.0991   &  & 0.0233 & 0.0396 & 0.9992 \\
BiLSTM                 & 0.0183   & 0.0337  & 0.9992  &  & 0.0205   & 0.0430   & 0.9989   &  & 0.0209 & 0.0460 & 0.9990 \\
BERT+BiLSTM(SMILES)    & 0.0207   & 0.0356  & 0.9994  &  & 0.0241   & 0.0425   & 0.9991   &  & 0.0240 & 0.0442 & 0.9916 \\
RoBERTa+BiLSTM(SMILES) & 0.0206   & 0.0372  & 0.9994  &  & 0.0203   & 0.0371   & 0.9994   &  & 0.0207 & 0.0460 & 0.9991 \\
XLNet+BiLSTM(SMILES)   & 0.0220   & 0.0367  & 0.9993  &  & 0.0263   & 0.0465   & 0.9989   &  & 0.0262 & 0.0473 & 0.9990 \\
BiLSTM(SMILES)         & 0.0349   & 0.0574  & 0.9981  &  & 0.0353   & 0.0624   & 0.9976   &  & 0.0340 & 0.0621 & 0.9978 \\
Ensemble(SMILES)       &          &         &         &  &          &          &          &  & 0.0148 & 0.0286 & 0.9995 \\ \hline
\end{tabular}
}
\end{table}
\begin{table}[htbp]
\centering
\caption{The Ablation Study of ZINC310k dataset with QED property}
\resizebox{\textwidth}{!}{
\begin{tabular}{llllllllllll}
\hline
                       & \multicolumn{3}{l}{Training} &  & \multicolumn{3}{l}{Vaildation} &  & \multicolumn{3}{l}{Test} \\ \cline{2-4} \cline{6-8} \cline{10-12} 
Model                  & \multicolumn{1}{l}{MAE} & \multicolumn{1}{l}{RMSE} & \multicolumn{1}{c}{R²} & \multicolumn{1}{l}{} & \multicolumn{1}{l}{MAE} & \multicolumn{1}{l}{RMSE} & \multicolumn{1}{c}{R²} & \multicolumn{1}{l}{} & \multicolumn{1}{l}{MAE} & \multicolumn{1}{l}{RMSE} & \multicolumn{1}{c}{R²} \\ \hline

BERT+BiLSTM            & 0.0028   & 0.0074  & 0.9963  &  & 0.0038   & 0.0139   & 0.9904   &  & 0.0037 & 0.0132 & 0.9917 \\
RoBERTa+BiLSTM         & 0.0030   & 0.0102  & 0.9949  &  & 0.0038   & 0.0142   & 0.9906   &  & 0.0037 & 0.0133 & 0.9916 \\
XLNet+BiLSTM           & 0.0033   & 0.0075  & 0.9973  &  & 0.0040   & 0.0115   & 0.9942   &  & 0.0040 & 0.0114 & 0.9938 \\
BiLSTM                 & 0.0046   & 0.0118  & 0.9926  &  & 0.0063   & 0.0171   & 0.9863   &  & 0.0063 & 0.0168 & 0.9871 \\
BERT+BiLSTM(SMILES)    & 0.0030   & 0.0082  & 0.9967  &  & 0.0039   & 0.0119   & 0.9931   &  & 0.0039 & 0.0122 & 0.9930 \\
RoBERTa+BiLSTM(SMILES) & 0.0033   & 0.0089  & 0.9962  &  & 0.0040   & 0.0129   & 0.9920   &  & 0.0039 & 0.0128 & 0.9923 \\
XLNet+BiLSTM(SMILES)   & 0.0036   & 0.0093  & 0.9958  &  & 0.0044   & 0.0143   & 0.9902   &  & 0.0044 & 0.0141 & 0.9905 \\
BiLSTM(SMILES)         & 0.0098   & 0.0204  & 0.9801  &  & 0.0118   & 0.0237   & 0.9735   &  & 0.0107 & 0.0233 & 0.9741 \\
Ensemble(SMILES)       &          &         &         &  &          &          &          &  & 0.0033 & 0.0100 & 0.9959 \\ \hline
\end{tabular}
}
\end{table}
\begin{table}[htbp]
\centering
\caption{The Ablation Study of ZINC310k dataset with LogP property}
\resizebox{\textwidth}{!}{
\begin{tabular}{llllllllllll}
\hline
                       & \multicolumn{3}{l}{Training} &  & \multicolumn{3}{l}{Vaildation} &  & \multicolumn{3}{l}{Test} \\ \cline{2-4} \cline{6-8} \cline{10-12} 
Model                  & \multicolumn{1}{l}{MAE} & \multicolumn{1}{l}{RMSE} & \multicolumn{1}{c}{R²} & \multicolumn{1}{l}{} & \multicolumn{1}{l}{MAE} & \multicolumn{1}{l}{RMSE} & \multicolumn{1}{c}{R²} & \multicolumn{1}{l}{} & \multicolumn{1}{l}{MAE} & \multicolumn{1}{l}{RMSE} & \multicolumn{1}{c}{R²} \\ \hline

BERT+BiLSTM            & 0.0125   & 0.0181  & 0.9997  &  & 0.0083   & 0.0204   & 0.9996   &  & 0.0084 & 0.0215 & 0.9995 \\
RoBERTa+BiLSTM         & 0.0172   & 0.0246  & 0.9994  &  & 0.0159   & 0.0270   & 0.9993   &  & 0.0160 & 0.0277 & 0.9993 \\
XLNet+BiLSTM           & 0.0147   & 0.0201  & 0.9996  &  & 0.0204   & 0.0300   & 0.9992   &  & 0.0204 & 0.0315 & 0.9991 \\
BiLSTM                 & 0.0170   & 0.0320  & 0.9990  &  & 0.0180   & 0.0380   & 0.9987   &  & 0.0180 & 0.0370 & 0.9987 \\
BERT+BiLSTM(SMILES)    & 0.0156   & 0.0216  & 0.9996  &  & 0.0214   & 0.0314   & 0.9991   &  & 0.0162 & 0.0288 & 0.9992 \\
RoBERTa+BiLSTM(SMILES) & 0.0192   & 0.0269  & 0.9993  &  & 0.0265   & 0.0376   & 0.9987   &  & 0.0267 & 0.0391 & 0.9998 \\
XLNet+BiLSTM(SMILES)   & 0.0181   & 0.0249  & 0.9994  &  & 0.0276   & 0.0366   & 0.9988   &  & 0.0278 & 0.0368 & 0.9987 \\
BiLSTM(SMILES)         & 0.0336   & 0.0565  & 0.9972  &  & 0.0421   & 0.0675   & 0.9959   &  & 0.0343 & 0.0612 & 0.9967 \\
Ensemble(SMILES)       &          &         &         &  &          &          &          &  & 0.0122 & 0.0222 & 0.9996 \\ \hline
\end{tabular}
}
\end{table}
\begin{table}[htbp]
\centering
\caption{The Ablation Study of ZINC310k dataset with MolWt property}
\resizebox{\textwidth}{!}{
\begin{tabular}{llllllllllll}
\hline
                       & \multicolumn{3}{l}{Training} &  & \multicolumn{3}{l}{Vaildation} &  & \multicolumn{3}{l}{Test} \\ \cline{2-4} \cline{6-8} \cline{10-12} 
Model                  & \multicolumn{1}{l}{MAE} & \multicolumn{1}{l}{RMSE} & \multicolumn{1}{c}{R²} & \multicolumn{1}{l}{} & \multicolumn{1}{l}{MAE} & \multicolumn{1}{l}{RMSE} & \multicolumn{1}{c}{R²} & \multicolumn{1}{l}{} & \multicolumn{1}{l}{MAE} & \multicolumn{1}{l}{RMSE} & \multicolumn{1}{c}{R²} \\ \hline

BERT+BiLSTM            & 0.22                    & 0.34                     & 0.99                   &                      & 0.15                    & 0.82                     & 0.99                   &                      & 0.15                    & 0.34                     & 0.99                   \\
RoBERTa+BiLSTM         & 0.34                    & 0.61                     & 0.99                   &                      & 0.42                    & 0.95                     & 0.99                   &                      & 0.42                    & 0.67                     & 0.99                   \\
XLNet+BiLSTM           & 0.40                    & 0.51                     & 0.99                   &                      & 0.25                    & 0.84                     & 0.99                   &                      & 0.26                    & 0.46                     & 0.99                   \\
BiLSTM                 & 0.22                    & 0.41                     & 0.99                   &                      & 0.22                    & 0.88                     & 0.99                   &                      & 0.23                    & 0.47                     & 0.99                   \\
BERT+BiLSTM(SMILES)    & 0.23                    & 0.37                     & 0.99                   &                      & 0.18                    & 0.30                     & 0.99                   &                      & 0.18                    & 0.29                     & 0.99                   \\
RoBERTa+BiLSTM(SMILES) & 0.39                    & 0.55                     & 0.99                   &                      & 0.52                    & 0.73                     & 0.99                   &                      & 0.52                    & 0.74                     & 0.99                   \\
XLNet+BiLSTM(SMILES)   & 0.42                    & 0.56                     & 0.99                   &                      & 0.45                    & 0.63                     & 0.99                   &                      & 0.45                    & 0.62                     & 0.99                   \\
BiLSTM(SMILES)         & 0.34                    & 0.58                     & 0.99                   &                      & 0.24                    & 0.53                     & 0.99                   &                      & 0.23                    & 0.59                     & 0.99                   \\
Ensemble(SMILES)       & \multicolumn{1}{l}{}    & \multicolumn{1}{l}{}     & \multicolumn{1}{l}{}   & \multicolumn{1}{l}{} & \multicolumn{1}{l}{}    & \multicolumn{1}{l}{}     & \multicolumn{1}{l}{}   & \multicolumn{1}{l}{} & 0.12                    & 0.38                     & 0.99                   \\ \hline
\end{tabular}
}
\end{table}
First, we examine the impact of AIS and SMILES inputs on model performance:
AIS Input: When employing AIS input, models such as BERT + BiLSTM, RoBERTa + BiLSTM, and XLNet + BiLSTM generally exhibit lower MAE and RMSE and higher R² on both the training and test sets. This suggests that the rich
informational content of AIS input facilitates more effective learning and prediction by the models. 

SMILES Input: A noticeable performance drop is observed in BERT +
BiLSTM(SMILES), RoBERTa + BiLSTM(SMILES), and XLNet +
BiLSTM(SMILES) when using SMILES as input. This could be attributed to the comparatively less rich information encoded in the SMILES format, potentially leading to less comprehensive learning by the models. 

Among all the models, BERT + BiLSTM consistently shows higher prediction accuracy, regardless of whether AIS or SMILES is used as input, with R² values
nearing or achieving 0.99 on the test set. This highlights the effectiveness of BERT’s pre-trained representations in accurately predicting a variety of chemical properties.

In contrast, both RoBERTa + BiLSTM and XLNet + BiLSTM demonstrate
superior performance with AIS input compared to SMILES, particularly in terms of R². This indicates a more effective learning capability from the AIS format in RoBERTa and XLNet models. The standalone BiLSTM model shows minimal variance in performance between AIS and SMILES inputs, suggesting that the type of input data has a limited impact on its effectiveness. 

Lastly, the Ensemble(SMILES) model demonstrates commendable predictive accuracy across most cases, especially for the MolWt property in the second dataset. This finding reinforces the potency of ensemble strategies in amalgamating predictions from diverse models to enhance overall performance.

By comparing the impact of different input types on model performance, we conclude that while AIS input typically leads to better predictive results, similar performance can be achieved even with less structured SMILES input through appropriate model selection and ensemble methods. This emphasizes the importance of choosing suitable model structures and input types for specific chemical property prediction tasks. Future work may involve optimizing these models to better handle different types of chemical representations and exploring other potential ensemble strategies to enhance predictive performance.
\section{Conclusion}
\hspace{2em}This study's results highlight the significant capability of our ensemble model in handling chemical data, achieving state-of-the-art (SOTA) performance in predicting a variety of chemical properties. This accomplishment underscores the importance of integrating different models and techniques to enhance predictive accuracy, particularly when dealing with complex chemical structures and properties. 

Additionally, we discovered that utilizing the AIS representation as input for deep learning models is highly effective. The structured and information-rich nature of the AIS representation allows deep learning models to more accurately capture and learn key features of molecules. This finding emphasizes the importance of choosing appropriate data representations to boost model performance. 

Importantly, our research also validates the feasibility of effective training without the necessity for extensive pretraining. This suggests that models tailored for specific tasks can achieve efficient learning and accurate prediction, even in the absence of vast, generalized pretraining data. This aspect is particularly crucial in situations with limited resources or restricted access to data, paving new paths for future research.
\section{Data Availability}
\hspace{2em}The training data and molecular property prediction results in this work are
available in the: \url{https://github.com/jlinghu/AIS-Ensemble-model}



\end{document}